\RequirePackage{fix-cm}

\documentclass[twocolumn]{svjour3}          
\smartqed  
\usepackage{graphicx}
\usepackage{hyphenat}
\usepackage{amsmath}
\usepackage{multirow}
\usepackage{multicol}
\usepackage[breaklinks=true]{hyperref}
\usepackage{breakcites}
\begin{document}

\title{Environment reconstruction on depth images using Generative Adversarial Networks
}


\author{Lucas P. N. Matias         \and
        Jefferson R. Souza         \and
        Denis F. Wolf
}


\institute{Lucas P. N. Matias \at
              Institute of Mathematic and Computer Science, University of S\~ao Paulo(USP), S\~ao Carlos, Brazil \\
              \email{lucas.matias@usp.br}           
           \and
           Jefferson R. Souza \at
              Computer Science College, Federal University of Uberl\^andia(UFU), Uberl\^andia, Brazil\\
              \email{jrsouza@ufu.br}
            \and
            Denis F. Wolf \at
              Institute of Mathematic and Computer Science, University of S\~ao Paulo(USP), S\~ao Carlos, Brazil \\
              \email{denis@icmc.usp.br}   
}

\date{}

\maketitle

\begin{abstract}
Robust perception systems are essential for autonomous vehicle safety. To navigate in a complex urban environment, it is necessary precise sensors with reliable data. The task of understanding the surroun\-dings is hard by itself; for intelligent vehicles, it is even more critical due to the high speed in which the vehicle navigates. To successfully navigate in an urban environment, the perception system must quickly receive, process, and execute an action to guarantee both passenger and pedestrian safety. Stereo cameras collect environment information at many levels, e.g., depth, color, texture, shape, which guarantee ample knowledge about the sur\-roun\-dings. Even so, when compared to human, computational methods lack the ability to deal with missing information, i.e., occlusions. For many per\-cep\-tion tasks, this lack of data can be a hindrance due to the environment incomplete information. In this paper, we address this problem and discuss recent methods to deal with occluded areas inference. We then introduce a loss function focused on disparity and environment depth data reconstruction, and a Generative Adversarial Network (GAN) architecture able to deal with occluded information inference. Our results present a coherent reconstruction on depth maps, estimating regions occluded by different obstacles. Our final contribution is a loss function focused on disparity data and a GAN able to extract depth features and estimate depth data by inpainting disparity images.
\end{abstract}

\section{Introduction}
\label{intro}

\begin{figure} 
\centering
\includegraphics[width=0.48\textwidth]{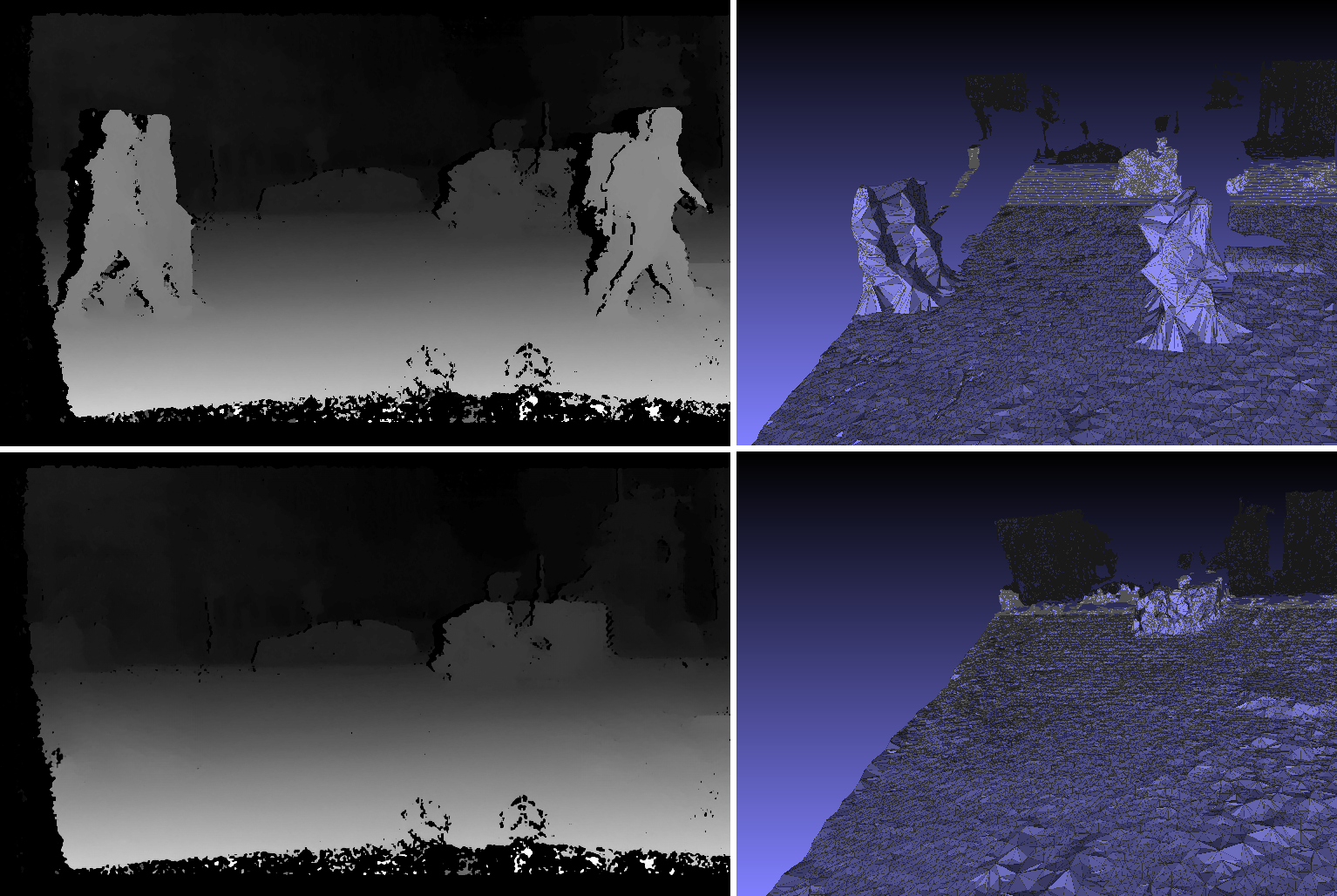}
\caption{Environment reconstruction from our methodology on disparity images and 3D mesh generated from the point cloud.}
\label{fig:intro_res}
\end{figure}

Intelligent Transportation Systems (ITS) have been wi\-dely studied to develop an autonomous vehicle able to na\-vi\-gate in a complex urban environment. To do so, intelligent vehicles depend on robust sensors to lo\-ca\-li\-ze themselves and interpret their surroundings. Precise perception systems are necessary for the decision-making algorithms; their confidence relies on accurate depth and visual information. Stereo cameras provide both information in a single frame, by capturing ima\-ges from multiple lenses and matching images features \cite{10.1007/978-3-642-19315-6_3}\cite{5539797}\cite{BMVC.25.14}. Based on matched features differences, a disparity value is calculated, then depth information can be estimated using the disparity values and camera parameters. Thereat, from stereo cameras it is possible to calculate depth data from RGB pixels, abstracting sensors synchronization problem.

Stereo camera collected data provides dense environment information to the perception system, which improves the knowledge extraction. However, in some cases, this data may not provide an ample environment insight. Dynamic obstacles moving through the scene can omit structural information, generating gaps of unknown depth data. For some tasks this missing information can be a problem, e.g. curb detection \cite{7225747}, mapping \cite{Yang2017Semantic3O}, instance segmentation \cite{DBLP:journals/corr/KhorevaBHHS16}, pedestrian intention detection \cite{8569415}. The human brain can deduce the missing data based on environmental context and previous experiences. For computational methods, it is a complex task to infer occluded depth information due to data sensitivity. By reconstructing these occluded are\-as, we would upsample the environment information, supporting many perception tasks and increasing their accuracy since it would become possible also to process and classify occluded regions.

Image reconstruction is already a vast research field, however, the majority of works are focused on RGB ima\-ges. Those works focus on a visual reconstruction, estimating color information on the target area based on neighborhood pixels. In a first look, those methods seem to be easily applicable to depth maps since disparity data encode depth information in a one-channel image. However, disparity images are too sensible, highlighting noises when a point cloud is reconstructed from the depth map. For a satisfactory depth reconstruction, a coherent and regular structure estimation is necessary. The reconstructed area should follow the real related structure distribution, diminishing the transition noise between real and generated areas (Fig. \ref{fig:intro_res}).

Convolutional Neural Networks (CNNs) have been extensively applied to image processing tasks, achie\-ving notable results \cite{8286426}. Goodfellow et al. \cite{NIPS2014_5423} introduced the Generative Adversarial Network architecture, which consists of two CNNs fighting to improve each other. This architecture can generate data very similar to the training dataset, which could be applied to image inpainting. GANs have been already proposed to deal with image reconstruction and inpainting, achieving notorious results on RGB images. Yet, the reconstructed area noises directly affects the tridimensional environment representation. The main problem with reconstructing depth maps are the data sensitivity and the few features to be extracted from those images. Withal, since disparity images encode depth information, some specific depth features could be considered. Based on that, one hypothesis arises: can we use depth data features to improve the GANs disparity data generation?

This work is divided into five sections: in Section \ref{intro} we discuss the gap on depth image inpainting and the motivations of our work, in Section \ref{related-approaches}, we present the recent efforts in image inpainting and depth completion research fields, and their flaws when applied to depth maps inpainting. Section \ref{methodology} presents an overview of disparity data aspects and our approach to deal with depth inpainting, in Section \ref{results} we discuss different metrics and their influence in GANs evaluation, then we present our qualitative and quantitative results compared with state-of-art approaches. Lastly, Section \ref{conclusion} summarizes our main contributions and future works.

\section{Related Approaches}
\label{related-approaches}

Image inpainting consists of an image reconstruction method that estimates image target regions \cite{6678248}. Those methods can also be applied to remove objects from the scene by reconstructing the area where the object once was. Inpaint depth maps means to estimate depth data; by removing an object, the reconstructed region is a guess of the area once occluded. However, deal with disparity data is more complicated than that since disparity images behavior is quite different from RGB ima\-ges. Depth completion studies deal with disparity data estimation by focusing on those unique characteristics. Study recent inpainting and depth completion methods can give an overview of recent efforts and methodologies to deal with image inpainting and depth maps data manipulation. With the knowledge about both fields, we can understand the gap in both research areas, and try to find a way to define a methodology for depth inpainting.

Criminisi et al. \cite{1323101} introduce one of the most re\-le\-vant inpainting algorithms. This method is based on an iterative patch analysis. The target area boundary is defined and divided into patches, at each iteration, is calculated the boundary gradients, and the unknown patch color is defined. At each iteration, known areas guide the estimation of unknown region patch by patch, growing the image reconstructed information. In the end, the region masked to be reconstructed is estimated. Most approaches \cite{Arias2011} proposes a similar method also based on iterative image patches analysis and growth. Those methods achieve a satisfactory image reconstruction on RGB images. However, they request many time to reconstruct large areas.

Some approaches \cite{5597710} \cite{7471866} deal with depth inpain\-ting. Those methods are based on the same iterative patch analysis. Since disparity images have only one image channel, the reconstruction process seems easier. However, due to the disparity images lack of features, those methods depend on many preliminary assumptions about the environment. They can only reconstruct small areas and, in many cases, only on well-controlled scenarios, e.g. indoor environments. In the context of autonomous vehicles, those methods could not be applied due to the high processing time and the environmental limitations in which they were proposed.

More recent methods tries to withdraw the proces\-sing time problem by using CNNs to inpainting images \cite{DBLP:journals/corr/PathakKDDE16} \cite{IizukaSIGGRAPH2017} \cite{yu2018generative}. Some of those works focus on GAN architectures to reconstruct the input image. Iizuka et al. \cite{IizukaSIGGRAPH2017} introduces a GAN network to deal with inpainting and object removal. This architecture divides the discriminator network on two, a local and a global discriminator. The global discriminator evaluates the whole reconstructed image and the estimation cohe\-rence. The local discriminator evaluates only the reconstructed area consistency. Yu et al. \cite{yu2018generative} proposes a post-processing architecture to improve this network result. A post-processing network receives as input the \cite{IizukaSIGGRAPH2017} network result, where the image real and gene\-rated regions are divided. With a contextual attention branch, real and generated images patches are analyzed and matched. After defining the image patches pairs, network tries to propagate real image characteristics to generated patches and their neighborhood, improving first network coarse result. Those approaches can achieve a result faster and do not depend on controlled environments. They are mainly applied to RGB images.

Based on the inpainting GAN architectures, some approaches reconstruct occluded information from semantic segmented images in order to estimate the road semantic layout \cite{DBLP:journals/corr/abs-1805-03356} \cite{DBLP:journals/corr/abs-1805-11746}. Those methods are also applied to outdoor urban scenarios. However, they reconstruct only semantic environment information. Those images have a small set of pixel colors to classify each pixel as one semantic class. Due to that, this task is simpler than RGB image inpainting and also less complex than the disparity reconstruction task.

Some methods are focused on estimating disparity image shadowed noisy areas generated by the lack of features during the stereo matching process. In that case, unknown disparity data are estimated, but they are based on known RGB information about the target area. However, different from inpainting approaches, those depth completion methods uses Conditional Random Fields (CRF) models in order to define a dispa\-rity value \cite{8460549} \cite{7533131} \cite{Buyssens2016DepthGuidedDI} \cite{7900062} \cite{7954738}. An objective function is defined, and then this function is minimized or maximized. To define the objective function, the known disparity data, and the RGB information associated with the unknown area are used to guide the disparity pixel estimation. In some cases, CNNs are used to extract features from RGB observable pixels - e.g. semantic information, surface normal vectors - improving the CRFs models \cite{dur22375} \cite{zhang2018deepdepth}. A temporal approach can also be used to estimate this disparity data \cite{7760974} \cite{Shen2013LayerDD}. In that case, consecutive frames are processed, and, based on the camera motion and previously calculated disparity data, new captured frames are completed using past environment depth information. 

Another depth reconstruction network is proposed in \cite{Uhrig2017SparsityIC}. This paper deal with depth upsample on LiDAR-based disparity images. Based on a single LiDAR disparity frame the CNN estimates the unknown depth information. This is a different kind of depth completion, which transforms LiDAR sensors sparse depth maps in a dense disparity image.

Inpainting methods can achieve a coherent visual data reconstruction. Although, when dealing with depth data, it is necessary a smooth and precise estimation due to data sensitivity. Some recent approaches use CNNs to deal with image inpainting, however even withdrawing the processing time problem, the noise reconstruction remains. Works focused on depth data are dependent on hard assumptions about the environment. Methodologies which does not depend on those assumptions estimate disparity data based on known RGB features related to the inpainted area. When dealing with occluded area estimation, we would not have information about area to be reconstructed. To estimate space behind an object, neither disparity nor RGB data associated with area available, so it would be useful to achieve an estimation by only using neighborhood data.

\section{Proposed Methodology}
\label{methodology}

\begin{figure}
\centering
\includegraphics[width=0.45\textwidth]{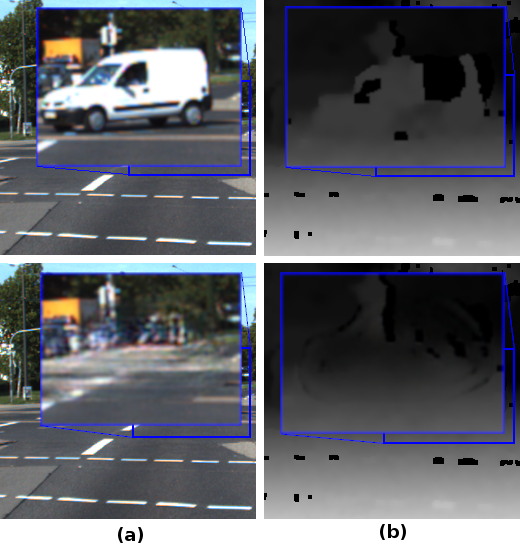}
\caption{Comparison between the Contextual Attention network applied to RGB and disparity images. (a) the RGB reconstruction without noticeable noise; (b) depth map reconstruction with noise artifacts generated on the transition between real and generated parts.}
\label{fig:rgb_disp}
\end{figure}

Yu et al. \cite{yu2018generative} proposes a GAN architecture to inpaint images achieving state-of-art results. Fig. \ref{fig:rgb_disp} displays this network result on RGB and disparity images. On RGB images, the network achieves a coherent and realistic reconstruction. However, on the disparity image, it is possible to notice a color variation inconsistency and noises in the reconstructed region. This inconsistency represents uneven reconstructed surfaces, those irre\-gularities are a problem on the point cloud generation, since they become artifacts on the tridimensional environment. When dealing with depth data, the local neighborhood has to be evaluated to minimize the irregularities on the generated surfaces, and maintain their depth continuity.

\begin{figure*}
\centering
\includegraphics[width=0.95\textwidth]{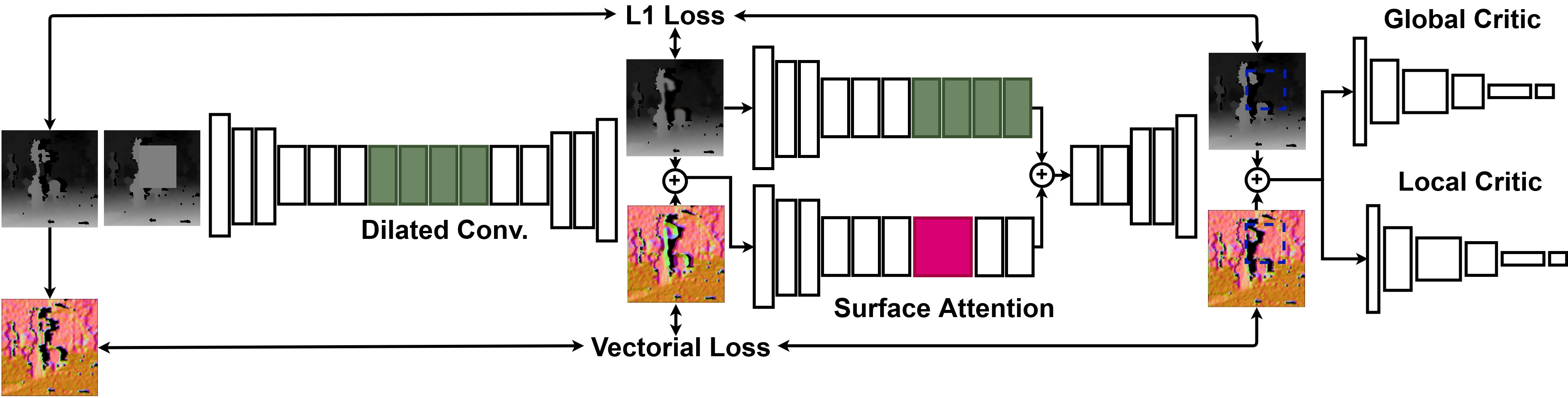}
\caption{Our proposed architecture, adding surface features to be evaluated by the contextual attention branch, loss function, and network discriminator.}
\label{fig:our_arch}
\end{figure*}

We propose a loss function oriented to disparity data, and modifications on \cite{yu2018generative} network architecture, focusing the context extraction and data evaluation on specific depth features (Fig. \ref{fig:our_arch}). With these adaptations, we want to improve the network disparity data, redu\-cing noise and uneven surface generation. With smoother disparity data, we can use our network to inpainting disparity images, estimating occluded depth information.

\subsection{Surface Features}

\cite{zhang2018deepdepth} uses a CRF model for disparity data estimation. To guide the estimation, firstly, they use a CNN to extract normal surface vectors from RGB images. The result is a coherent and regular estimation of the mis\-sing disparity data. However, they depend on known RGB information related to the unknown disparity region. Yet, it is possible to extract normal surface vectors from depth data. \cite{BESL198633} \cite{7335535} \cite{sripaper} \cite{harms2014accuracy} proposes diffe\-rent approaches to estimate surface normal vectors from disparity images. Those methods calculate the depth values from disparity data and then estimate normal surface vectors. The problem of using real depth values is that the specific camera parameters are necessary to calculate depth from disparity data. For different ca\-me\-ras, those parameters can change, as well as the image size. To apply that to a CNN, it would be problematic and limiting. Since disparity images encode depth value in a one-channel image, the dis\-pa\-rity data variation is proportional to the depth variation. We can try to calculate surface normal vectors from disparity va\-lues only. This would be just an approximation of the real normal vectors; due to the proportionality, this estimation would be consistent with the surface variation. 

\begin{figure}
\centering
\includegraphics[width=0.45\textwidth]{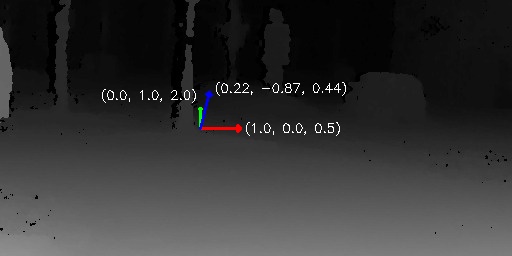}
\caption{Vector normal to a pixel calculated from image $x$ and $y$ axis disparity gradients.}
\label{fig:surf_disp}
\end{figure}

From definition, the cross product between two orthogonal vectors $\overrightarrow{v_1}$ and $\overrightarrow{v_2}$ is a vector $\overrightarrow{n}$ perpendicular to both vectors, i.e. $\overrightarrow{v_1} \times \overrightarrow{v_2} = \overrightarrow{n}$ where $\overrightarrow{n} \perp \overrightarrow{v_1}$ and $\overrightarrow{n} \perp \overrightarrow{v_2}$. Which means that, the resultant vector is normal to the plane where $\overrightarrow{v_1}$ and $\overrightarrow{v_2}$ belongs. The gradients on $x$ and $y$ image axis for a particular disparity pixel, give the disparity variation on that neighborhood. To calculate the vector normal to a pixel surface, we take two unitary vectors parallel to the $x$ and $y$ image axis and define their $z$ component as the $x$ and $y$ directional gradient value on that region. Thus we will have two vectors that are in a plane parallel to the surface in which the pixel lays on. The cross product between these two vectors is a resultant normal to the pixel surface (Fig. \ref{fig:surf_disp}). For a pixel $P_{i,j}$ Equation \ref{eq:norm} displays the formulation for the normal vector calculation.

\begin{equation}
    \label{eq:norm}
    \begin{aligned}
    P_{\Delta i} = \frac{P_{i+1,j} - P_{i-1,j}}{2}
    \\
    P_{\Delta j} = \frac{P_{i,j+1} - P_{i,j-1}}{2}
    \\
    \overrightarrow{v_{\Delta i}} = (1.0, 0.0, P_{\Delta i})
    \\
    \overrightarrow{v_{\Delta j}} = (0.0, 1.0, P_{\Delta j})
    \\
    \overrightarrow{v} = \overrightarrow{v_{\Delta i}} \times \overrightarrow{v_{\Delta j}}
    \\
    \overrightarrow{n} = \frac{\overrightarrow{v}}{\left\lVert\overrightarrow{v}\right\rVert}
    \end{aligned}
\end{equation}

\begin{figure}
\centering
\includegraphics[width=0.45\textwidth]{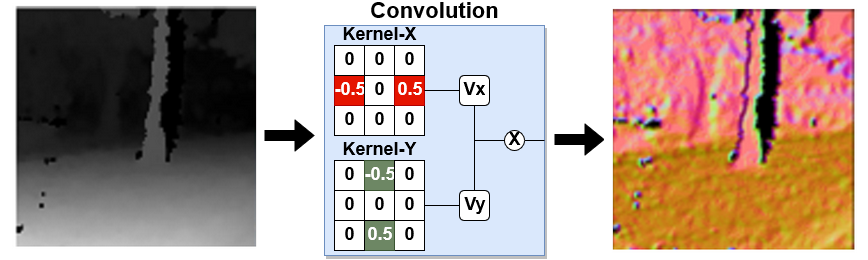}
\caption{Surface normal vectors image from depth maps after the surface convolution.}
\label{fig:kernel_norm}
\end{figure}

Using the image gradient only to calculate normal vectors for disparity pixels can give information about the surfaces of the structure, withdrawing the camera parameters limitation. Also, since we use the neighborhood pixels to calculate the normal vectors, we can model this calculation as a convolution (Fig. \ref{fig:kernel_norm}). This convolution could be added to the CNN training process without increasing the processing time significantly.

\subsection{Vectorial Loss}
The Contextual Attention architecture uses the Wasserstein GAN (WGAN) \cite{2017arXiv170107875A} as the loss function, which has a smoother gradient than the original GAN loss. Du\-ring the training, a gradient penalty is applied to avoid the gradient vanishing problem (WGAN-GP) \cite{2017arXiv170400028G}. Also, an L1 loss function is calculated to quantify the diffe\-rence between generated and ground truth image pi\-xels. Although, this pixel-wise loss evaluates only the local error, increasing the color variation around the pixel neighborhood. This high variation generates uneven surfaces on the disparity reconstructed data. Since disparity gradients are used to estimate a vector normal to the pixel surface, the difference between ground truth and generated surface normal vectors can quantify the pixel neighborhood error. By evaluating the reconstructed area normal vectors, it is possible to ve\-rify the generated surface consistency. This loss function would evaluate depth features, enhancing the network depth data processing.

By calculating the normal surface vectors from ground truth and generated disparity images, we can evaluate the vectorial error. With this Vectorial Loss, the GAN may enhance the structure's surface generation. By diminishing the generated surfaces irregularities and removing noisy artifacts, the reconstructed disparity image would be more realistic. Therefore, we maintain the original loss function with the WGAN-GP and L1 losses and add the Vectorial Loss as a third factor to be minimized. As the L1 loss function, we calculate the Vectorial error on the final and coarse results. Thus, we mi\-nimize the surfaces generation error on both networks:

\begin{equation}
    \label{eq:new_loss}
    \begin{aligned}
    V_l(X_{\overrightarrow{n}},Y_{\overrightarrow{n}}) = \sum_{1}^{i} \sum_{1}^{j} \mid x_{\overrightarrow{n}} - y_{\overrightarrow{n}} \mid
    \end{aligned}
\end{equation}

\begin{equation}
    \label{eq:new_loss}
    \begin{aligned}
    d_l = W(P_r, P_g) + GP(Y, m)
    \\
    g_l = \beta G(Y) + \phi L1(X,Y) + \alpha V_l(X_{\overrightarrow{n}},Y_{\overrightarrow{n}})
    \end{aligned}
\end{equation}

Where $d_l$ is the discriminator WGAN-GP loss and $g_l$ the generator loss function. $G(Y)$ is the WGAN-GP generator loss calculated for the generated image, $L1(X,Y)$ is the L1 loss between generated and ground truth images and $V_l(X_{\overrightarrow{n}},Y_{\overrightarrow{n}})$ is the Vectorial Loss between generated and ground truth surface normal vectors. $\beta$, $\phi$ and $\alpha$ are predefined parameters.

\subsection{Surface Attention branch}
The contextual attention branch proposed in \cite{yu2018generative} divides the first network coarse result into two classes, background, and foreground. The real image area is defined as the background while the reconstructed region defined as the foreground. Background patches are divided on $3x3$ patches and modeled as a convolution. With a cosine similarity calculation and a softmax function, a probability score is calculated between real patches and generated pixels.



To maintain the coherence between adjacent pixels, the calculated attention score is propagated to neighborhood pixels. This adjacent propagation is done by a left-right and top-down attention score shift on a kernel of size $k$. By summing the kernel attention scores, the patches inherit their adjacent pixels matched characteristics, maintaining the neighborhood coherence. The highest scored patch convolution is used to deconvolve the matched generated pixel and propagate the matched real characteristics. This branch extracts context information from the coarse reconstructed image and processes it on the second network deconvolution, improving the inpainting result. 


However, due to the lack of color features, disparity pixels only may do not have enough contextual information to guide the characteristics propagation. Match disparity pixel means to match only depth information, which does not give any context about the structure in which that pixel belongs. Without enough data, the attention score calculation may do not find the most similar patches, replicating characteristics not related to the surface in which the pixels lays. To improve the context extraction, we give more information to be processed by this branch. We calculate the normal surface vectors from the coarse result. Those normals are concatenated to the disparity image to be processed by the contextual attention branch. We extract relevant features about the scene reconstructed structures and improve the foreground and background regions comparison. Since this branch process disparity pixels and their normal vectors, surface information also will be propagated to the reconstructed disparity region on the deconvolution, enhancing generated depth structures.

\subsection{Surface Discrimination}
The basis for GAN is the two networks disputing at each iteration. The discriminator network is fundamental since it learns to distinguish between real and fake data, improving the generator result. On this architecture, we would provide the generator reconstructed disparity image to the discriminator. However, as discussed before, depth maps have low visual features to be extracted, which could lead the discriminator to focus on features not related to depth information.

Since the discriminator network can well divide both real and fake data, we want to use this ability to distinguish also between real and fake depth surfaces. Instead of giving only the disparity image to the discriminator network, we calculate the normal surface vectors from the output image and also provide this information to be evaluated. With this surface information associated with the depth map, the discriminator can relate both disparity and surface data to distinguish between real and fake depth images. Thus, we guarantee that the discriminator will look after depth features, evaluating the surface generated irregularities. After the convergence, we use the generator to reconstruct depth maps and withdraw the final surface normal vectors calculation step. With this approach, the GAN will be focused on depth data evaluation at every training step. At the convergence, the generated depth maps will have disparity and surface distributions close to the real data.

\section{Experimental Results}
\label{results}

To evaluate our network in the context of autonomous vehicles, stereo images from an urban scenario are ne\-cessary. The KITTI dataset \cite{Geiger2013IJRR} \cite{Geiger2012CVPR} is a benchmark on ITS research field. This dataset collects images from different scenarios around the city of Karlsruhe, Germany. However, this dataset has only the stereo RGB pair images. Since those images are collected in an outdoor environment, the sunlight can be an issue to the disparity matching algorithms, generating gaps of unknown disparity information. Due to that, we have used the CityScapes dataset \cite{Cordts2016Cityscapes} \cite{Cordts2015Cvprw}. This dataset collects images from many Germany cities, in distinct weather conditions for a more embracing data context. Those images are post-processed and manually selected, then semantic, and instance information is annotated, and disparity images are calculated. Due to the post-proces\-sing and manual image selection, the final disparity i\-mages are denser, diminishing the shadowed gaps. These less noisy data guarantee more information to be extracted and matched by the network on image reconstruction.

\begin{figure}
\centering
\includegraphics[width=0.45\textwidth]{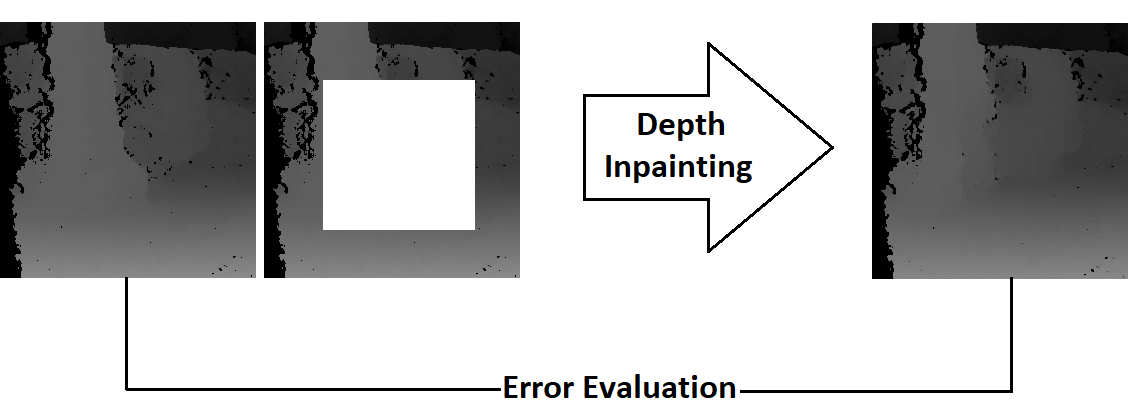}
\caption{Numerical network evaluation, where we crop the area to be reconstructed by the network, with the original removed data as the ground truth.}
\label{fig:eval_quant}
\end{figure}

To evaluate our proposed architecture, we compare the original state-of-art Contextual Attention network \cite{yu2018generative} with our approach. Since we introduce our Vectorial Loss and many adaptations around the original network architecture, we incrementally apply those mo\-difications, train the modified architecture, and evaluate their results. With this incremental evaluation, we can follow the network improvements, and evaluate our Vectorial Loss and each adaptation influence to the final network architecture. For a quantitative assessment, we use the validation set squared images with a random $128\times128$ crop around it to be reconstructed by our network. Then, we calculate the error between the reconstructed and ground truth images (Fig. \ref{fig:eval_quant}). For the qualitative evaluation, we use different sized i\-mages with specific objects masked to be removed in the scene, so we can evaluate how satisfactory is the occluded region estimation.

\begin{figure*}
\centering
\includegraphics[width=1.0\textwidth]{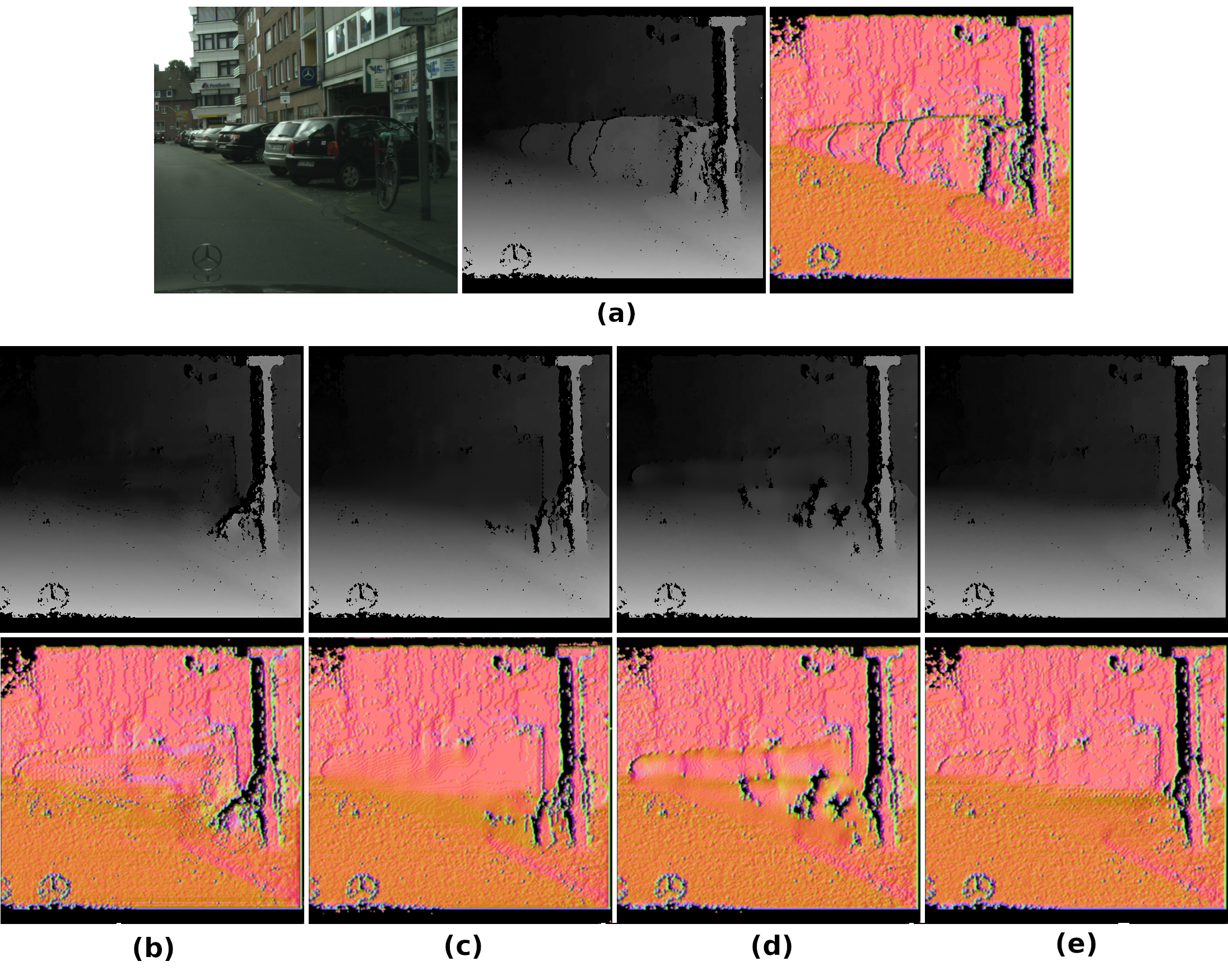}
\caption{Depth inpainting results from original Contextual Attention network \cite{yu2018generative} and our adaptations. (a) The original RGB and respective disparity and surface normals image; (b) the Contextual Attention network results; (c) results with the Vectorial Loss; (d) results with Vectorial Loss and Surface Attention adaptations; (e) results from our final proposed architecture.}
\label{fig:disp_surf}
\end{figure*}

Fig. \ref{fig:disp_surf} displays the results comparison between the original and the adapted architectures. The original Contextual Attention (Fig. \ref{fig:disp_surf}.b) network generates disparity data with a high variation around the neighborhoods, creating undesirable artifacts. Adding the Vectorial Loss helps the network to create smoother surfaces, removing the noisy artifacts. Even so, the characteristics of the structure are lost (Fig. \ref{fig:disp_surf}.c). The Surface Attention branch improves the structures reconstruction by propagating real surface information to the gene\-rated area. The network proposes different structures to reconstruct the image, yet they may do not correctly fit on the area where they are placed (Fig. \ref{fig:disp_surf}.d). When we look at the surface images, it is possible to notice that all those reconstructed data distributions are very far from real surface distributions. When we evaluate the generated surfaces on our proposed network architecture, the surface data distribution is improved, enhan\-cing their reconstruction (Fig. \ref{fig:disp_surf}.e). By surface discrimination, the network achieves a coherent reconstruction of disparity and surface images. Our proposed network is then able to remove the cars from the scene and estimate the environment behind, generating a smooth depth interpolation with consistent surfaces.

\begin{figure*}
\centering
\includegraphics[width=0.9\textwidth]{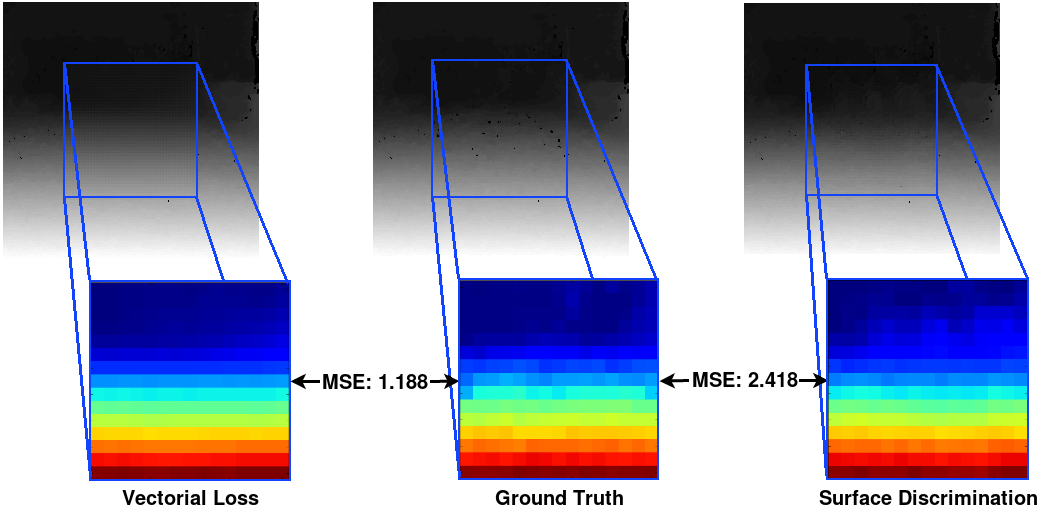}
\caption{Mean squared error from the network without and with Surface Discrimination.}
\label{fig:distrib_comp}
\end{figure*}

However, when we calculate the MSE and VE for the validation images, we notice that the quantitative evaluation does not agree with the same improvements. Table \ref{tab:pixel_err} shows the MSE and VE, where CA is the Contextual Attention original network, VL is our proposed Vectorial Loss, and SA is our Surface Attention branch adaptation. On this table, our final network result is worst than previous adaptations and the original network, even with a better image reconstruction. This evaluation does not seem correct, and this can be explained by the way that GANs works. GANs tries to replicate the training set data distribution, which means that given input data, an output data will be gene\-rated trying to imitate the real data distribution. Even with the L1 and Vectorial Loss, which calculates the error for each pixel, the GAN focus on recreating structures with the same aspect as the real ones. Without Surface Discrimination, the only metric used to evaluate the generated surfaces was the Vectorial Loss. By only analyzing the individual pixel error, the network tries to find an intermediary value that will best fit in that neighborhood, which explains the smooth aspect of the generated surfaces. When we start discriminating real and generated surfaces, the network understands that those smoother surfaces do not look real and propose surfaces with some subtle roughness, as the real ones. However, the regions where those imperfections are placed may do not match with the areas where they are placed on the ground truth image. Figure \ref{fig:distrib_comp} displays the reconstructed region comparison in our proposed network with and without the Surface Discrimination. Thus, the pixel-wise error metrics will increase, since the generated data distribution does not locally fit with the ground truth data. Yet, when analyzing the whole generated area data distribution, it is much closer to the ground truth.

\begin{table}[]
\caption{Mean-Squared Error and Vectorial Error calculated from each network adaptation and original Contextual Attention network.}
\label{tab:pixel_err}
\centering
\begin{tabular}{ccc}
\hline\noalign{\smallskip}
        & MSE            & VE         \\
\noalign{\smallskip}\hline\noalign{\smallskip}
CA \cite{yu2018generative}     & 189.078          & 0.1212          \\
CA \cite{yu2018generative} + VL & \textbf{181.870} & 0.0908          \\
SA + VL & 194.335          & \textbf{0.0907} \\
Ours    & 197.527          & 0.1054         
\end{tabular}
\end{table}

Depending on the application, pixel-wise metrics can give an erroneous performance evaluation of the GAN. As in our case, where depth data are too sensible to neighborhood variation, those metrics can not precisely evaluate how well the network replicates the training dataset. We propose the use of data distribution comparison metrics to evaluate network performance. By computing a histogram of the network image result, we can effectively compare the generated and real depth data distributions, avoiding local pixel evaluations.

\subsection{Distributions Distance}
Statistical distance metrics can quantify the simila\-rity between two data distributions by a distance or a percentage value. To evaluate our results, we use image histograms and the statistical distance metrics to quantify how well our network can imitate the training set data distribution. \cite{2017arXiv170107875A} introduce their WGAN loss function based on the Wasserstein distance metric. Some other distribution distance metrics are discussed on that work, e.g., Kullback-Liebler and Jensen-Shannon divergence. In addition to these, other metrics can also be used to compare images histograms. We can calculate the percentage of matching bins from both histograms by the histogram intersection \cite{10.1007/978-3-642-77225-2_13}. Also, the correlation between two histograms can be calculated, which quantify how much two data distributions are related. Using ground truth and generated images, we can apply those metrics to compute the similarity between real and ge\-nerated image data distributions.


We apply those metrics on the generated dispa\-rity image and their respective normal vectors. By using many distinct data distribution comparison metrics, we can better understand the network data generation. Those distance values give information about the global data generation and how the network deals with the environmental context. Pixel-wise metrics quantify only the individual pixel error, leading to a wrong interpretation of the network reconstruction. Those statistical distance metrics evaluate the image data distribution, ignoring local errors, and considering the whole image consistency.

\begin{table*}[]
\caption{Data distribution distances between ground truth and reconstructed images for each applied adaptation.}
\label{tab:stat_dist}
\begin{tabular}{ccccccccccc}
\hline\noalign{\smallskip}
\multirow{2}{*}{} & \multicolumn{2}{c}{Jensen-Shannon}  & \multicolumn{2}{c}{Kullback-Liebler} & \multicolumn{2}{c}{Wasserstein}     & \multicolumn{2}{c}{Hist. Intersection} & \multicolumn{2}{c}{Hist. Correlation} \\
                  & Depth            & Surface          & Depth             & Surface          & Depth            & Surface          & Depth                & Surface             & Depth               & Surface             \\
\noalign{\smallskip}\hline\noalign{\smallskip}
CA \cite{yu2018generative}                & 0.4635          & 0.3845          & 0.6979           & 0.6536          & 0.00257          & 0.0021          & 0.5085              & 0.5667             & 0.6135             & 0.4258             \\
CA \cite{yu2018generative} + VL           & 0.4749          & 0.3887          & 0.7428           & 0.6124          & \textbf{0.00240}          & 0.0015          & 0.4917              & 0.5520             & 0.5839             & 0.4727             \\
SA + VL           & \textbf{0.4550} & 0.3862          & \textbf{0.6681}  & 0.6066          & 0.00268          & \textbf{0.0015} & 0.5206              & 0.5557             & \textbf{0.6620}    & 0.4772             \\
Proposal              & 0.4556          & \textbf{0.3797} & 0.6726           & \textbf{0.5987} & 0.00255 & 0.0018          & \textbf{0.5232}     & \textbf{0.5660}    & 0.6583             & \textbf{0.5126}  
\end{tabular}
\end{table*}

Table \ref{tab:stat_dist} displays the results using the data distance metrics. On this table, it is possible to notice an incremental improvement as we add our proposed adaptations to the original network. This improvement follows what we see on the qualitative evaluation. The Vectorial Loss interfered with the disparity data generation, increasing the error; however, this loss function could improve the surface generation, reducing surface data distribution distance to the ground truth. When we add the Surface Attention to the network, we could improve both disparity and surface data. By analyzing the normal vectors, the contextual attention branch could match more consistent features and replicate real surface characteristics. The last row has the evaluation for our final proposed network, with Vectorial Loss, Surface Attention branch, and Surface Discrimination. Those results show a slight increase of disparity error on some metrics, yet this subtle increase occurs at the expense of a significant improvement on surface data generation. The improvement on the surface distribution is more relevant since this calculation is based on the neighborhood pixel evaluation, representing a coherent structure depth surface reconstruction. That behavior means that even estimating not such precise objects or structures distances, their surfaces will be consistent and realistic. 

\begin{figure*}
\centering
\includegraphics[width=1.0\textwidth]{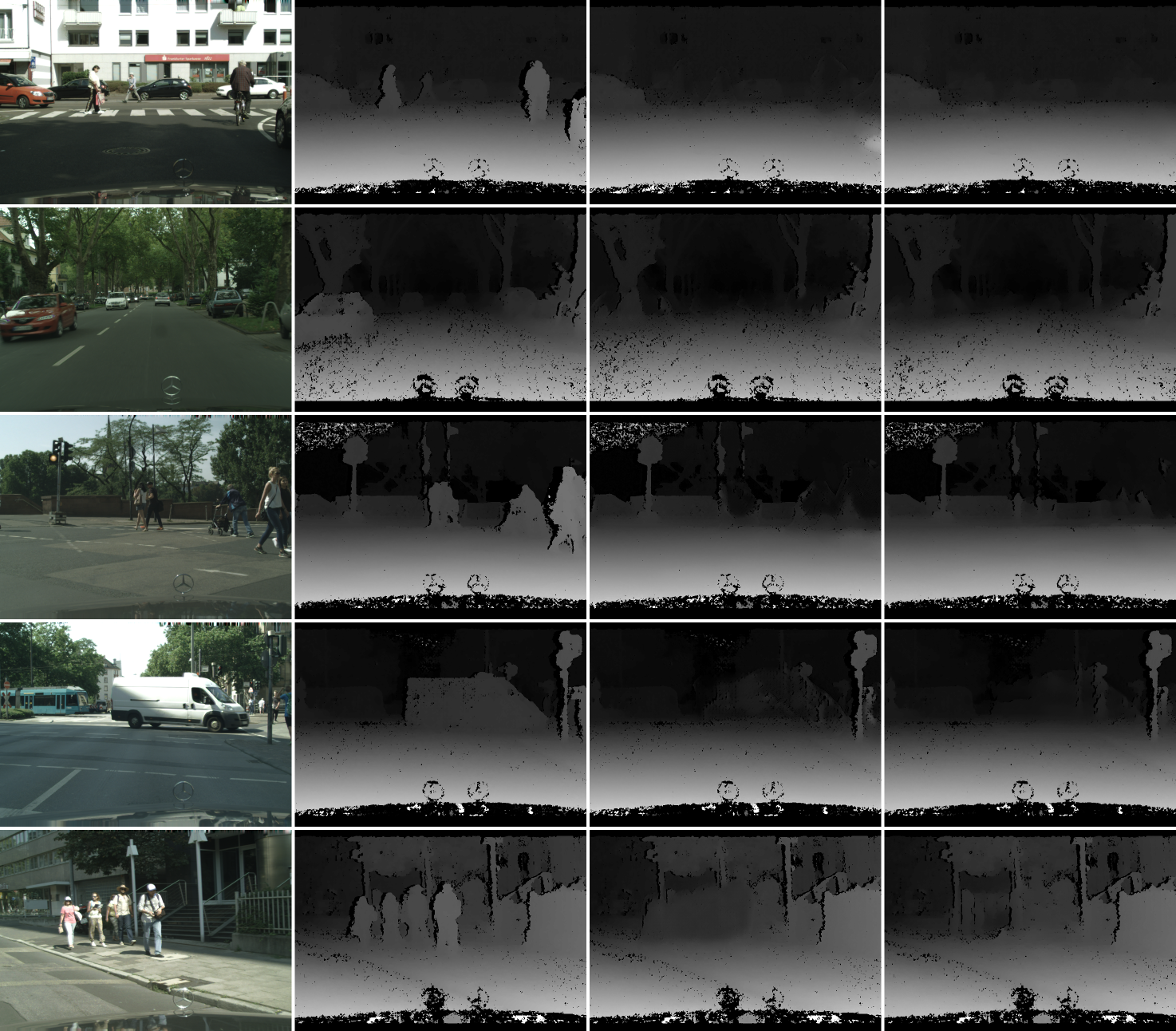}
\caption{Comparison between Contextual Attention network \cite{yu2018generative} depth reconstruction and ours. First column the RGB image; Second column the original disparity image; Third column the Contextual Attention reconstruction of the area behind the target objects; Last column our reconstruction.}
\label{fig:res_compare}
\end{figure*}

Fig. \ref{fig:res_compare} compare our network inpainting with the ori\-ginal Contextual Attention network on many different scenes and removing different objects. It is possible to notice that our proposed architecture was able to understand the depth surfaces context and reconstruct the environment with consistent structures. Even when the Contextual Attention network estimates a smooth interpolation, our network can generate more realistic structures. Also, continuous surfaces - i.e., wall, ground, pole - have an even interpolation avoiding the noisy artifacts generation, proposing uniform structures.

\section{Conclusion}
\label{conclusion}

The Image Inpainting research field has achieved pro\-mis\-sory results. GAN based methodologies can recons\-truct image areas replicating real image context to the ge\-ne\-ra\-ted region. The challenge of applying those me\-thods on disparity images are the data sensitivity and few features to be extracted to guide the reconstruction. By studying recent Inpainting and Depth Completion methods, we were able to understand the Depth Inpainting task's main weaknesses. In this work, we propose a new loss function and a GAN architecture adaptation, based on depth maps surface normal vectors features. Using only disparity pixel neighborhood values, we estimate vectors normal to the pixel surface, obtaining depth-related characteristics. By quantifying the error and providing generated surface information, we point out to the network the surface consistency importance when estimating disparity values.

Our results display a coherent environment recons\-truction on depth maps for different scenes and recons\-tructed area sizes. The disparity data inference follows the characteristics of the real structure, generating uniform surfaces. Our network could achieve a con\-si\-de\-ra\-bly better disparity image reconstruction than state-of-art Inpainting networks. This Depth Inpainting network application analysis will be left for future works, from researchers who may need to estimate occluded depth information for perception tasks.

\begin{figure*}
\centering
\includegraphics[width=1.0\textwidth]{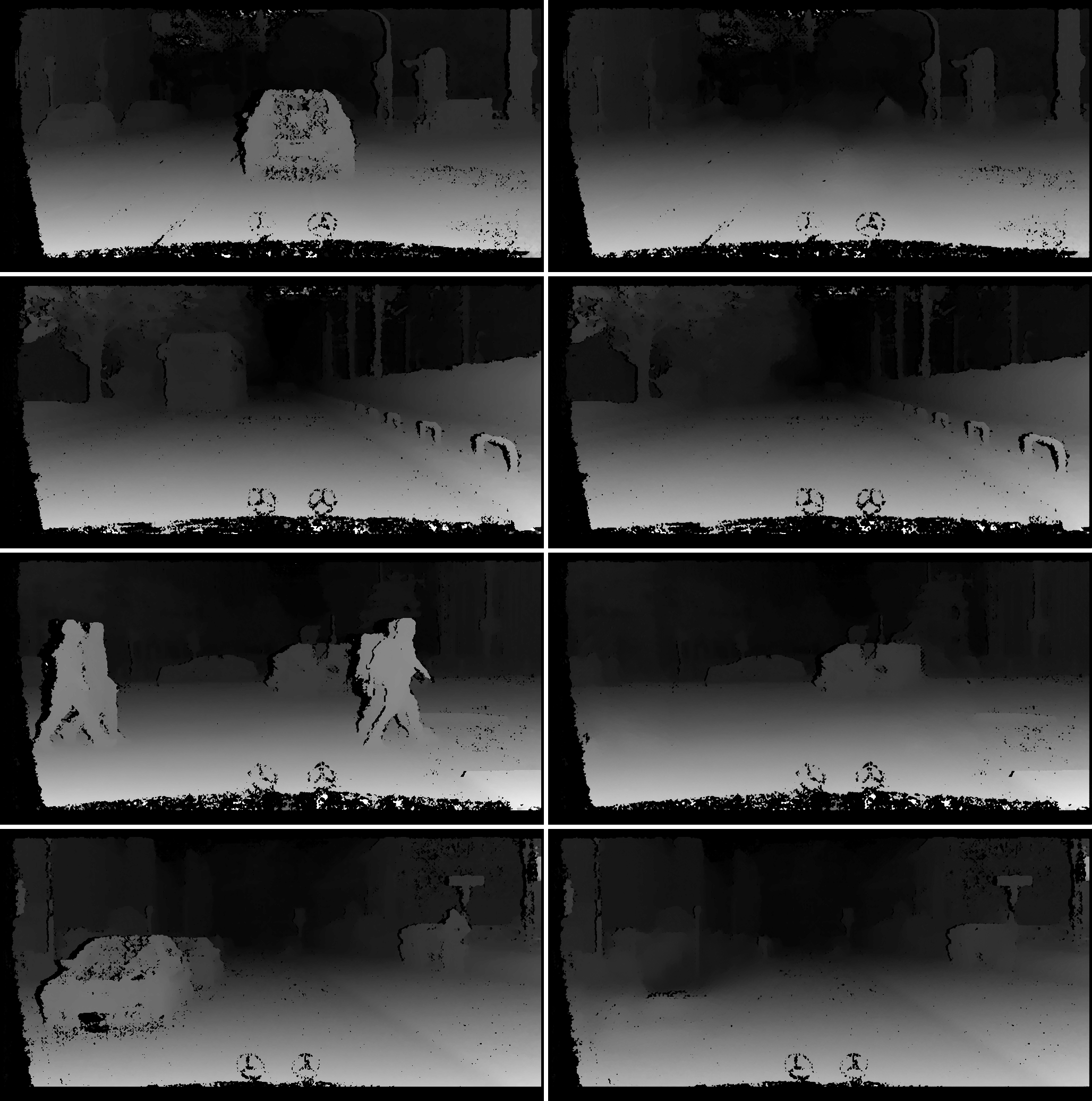}
\caption{More scenes reconstructed by our network. First column the original disparity image and on the second column some objects removed and the estimated occluded region.}
\label{fig:more_res}
\end{figure*}


%
%


%
%
\bibliographystyle{spmpsci}
\bibliography{References.bib}

\end{document}